\documentclass[conference]{IEEEtran}
\IEEEoverridecommandlockouts
\usepackage{cite}
\usepackage{amsmath,amssymb,amsfonts}
\usepackage{algorithmic}
\usepackage{graphicx}
\usepackage{textcomp}
\usepackage[table]{xcolor}
\def\BibTeX{{\rm B\kern-.05em{\sc i\kern-.025em b}\kern-.08em
    T\kern-.1667em\lower.7ex\hbox{E}\kern-.125emX}}

\usepackage{tcolorbox}
\usepackage{multirow}
\usepackage{svg}

\definecolor{custom_light_blue}{rgb}{0.85, 0.95, 1}
\definecolor{custom_light_pink}{rgb}{1, 0.85, 0.85}

\makeatletter
\newcommand{\linebreakand}{%
  \end{@IEEEauthorhalign}
  \hfill\mbox{}\par
  \mbox{}\hfill\begin{@IEEEauthorhalign}
}
\makeatother

\begin{document}

\title{Transformer Encoder and Multi--features Time2Vec for Financial Prediction\thanks{Supported by the University Excellence Fund of Eötvös Loránd University, Budapest, Hungary (ELTE). Nguyen Kim Hai Bui and Nguyen Duy Chien contributed equally to this work.}
}

% Long-format
\author{\IEEEauthorblockN{\bf Nguyen Kim Hai Bui\IEEEauthorrefmark{1}, Nguyen Duy Chien\IEEEauthorrefmark{1}, P\'{e}ter Kov\'{a}cs\IEEEauthorrefmark{2}, and Gerg\H{o} Bogn\'{a}r\IEEEauthorrefmark{2}}
\IEEEauthorblockA{\IEEEauthorrefmark{1}\textit{ELTE Eötvös Loránd University}}
\IEEEauthorblockA{\IEEEauthorrefmark{2}\textit{Department of Numerical Analysis, ELTE Eötvös Loránd University}}
\IEEEauthorblockA{\texttt{\{qmibhu,r0iftm,kovika\}@inf.elte.hu}, \texttt{bognargergo@staff.elte.hu}}}
 
\maketitle

\begin{abstract}
Financial prediction is a complex and challenging task of time series analysis and signal processing, expected to model both short-term fluctuations and long-term temporal dependencies. Transformers have remarkable success mostly in natural language processing using attention mechanism, which also influenced the time series community. The ability to capture both short and long-range dependencies helps to understand the financial market and to recognize price patterns, leading to successful applications of Transformers in stock prediction. Although, the previous research predominantly focuses on individual features and singular predictions, that limits the model’s ability to understand broader market trends. In reality, within sectors such as finance and technology, companies belonging to the same industry often exhibit correlated stock price movements.

In this paper, we develop a novel neural network architecture by integrating Time2Vec with the Encoder of the Transformer model. Based on the study of different markets, we propose a novel correlation feature selection method. Through a comprehensive fine-tuning of multiple hyperparameters, we conduct a comparative analysis of our results against benchmark models. We conclude that our method outperforms other state-of-the-art encoding methods such as positional encoding, and we also conclude that selecting correlation features enhance the accuracy of predicting multiple stock prices.

\end{abstract}

\begin{IEEEkeywords}
time series analysis, financial prediction, neural networks, feature engineering
\end{IEEEkeywords}

\section{Introduction}
Time series forecasting, especially in the financial sector, is a challenging, yet critically demanded and safety critical problem. Market prices are influenced by a multitude of factors, leading to complex investment and trading strategies that are required to model and manage multiple aspects \cite{pedersen2019efficiently}. In this paper, we focus on the particular problem of stock price prediction. The stock market can be viewed as a dynamic system, where stocks and further financial commodities (e.g. currencies, equities) are being traded. Traders generally prefer to purchase stocks of companies that show potential for future growth and avoid those expected to decline in value \cite{stockforecast}. {Companies within the same sector often exhibit similar correlations, as they tend to be affected similarly by macroeconomic events such as economic downturns or political policies}. The most commonly used methods for predicting stock returns are broadly categorized into fundamental analysis and technical analysis. Fundamental analysis aims to determine the intrinsic value of stocks based on internal and external economic indicators, that is particularly essential for investors focused on long-term trading. Technical analysis, being investigated in this paper, focuses on techniques to maximize the capital gain and decrease the loss, based on statistical analysis that is demanded to forecast the stock price \cite{indianstock}. Fast and accurate stock price forecasting is difficult, and this field is an active area of research where the best ways are yet to be found \cite{stockforecast}.

In principle, time series forecasting are expected to understand and model complex linear and non-linear interactions within historical data to predict the future. Traditional statistical approaches commonly adopt linear regression, exponential smoothing \cite{gardner1985forecasting}, and auto regression model \cite{SEMENOGLOU20211072}, see e.g. the well-known Auto Regressive Integrated Moving Average (ARIMA) or Seasonal ARIMA (SARIMA) methods. With the advances in deep learning, recent works are heavily invested in ensemble models and sequence-to-sequence modeling such as recurrent neural networks (RNNs), and long short-term memory (LSTM) \cite{6795963}, in particular. However, the primary drawback of these methods is that the RNN family struggles to capture extremely long-term dependencies \cite{DBLP:journals/corr/abs-1907-00235}. The recently popularized sequence-to-sequence model called Transformer \cite{Transformer} has achieved great success in natural language processing (NLP), especially in machine translation and large language models (LLMs), but also influenced other fields like signal processing and computer vision. Different from RNN-based models, transformers employ self-attention mechanism to learn the relationship among different positions globally. Motivated by the success of deep learning in image classification, natural language processing, and time series analysis, deep learning has become an emerging approach in the field of finance as well. Deep neural networks has the ability to learn low-dimensional representation extracted from originally high-dimensional input data. Machine and deep learning techniques have been found to be efficient in price forecasting and market trend prediction, see e.g. \cite{predictlSTM}, that support investors to anticipate future trends and optimize stock investments.

This work investigates stock price forecasting with deep learning, offering multiple novel contributions. First, we suggest a multi-feature selection method based on analysis of multiple correlating stock prices, then we propose a novel neural network architecture involving Time2Vec \cite{Time2vec} as encoding, combined with the Encoder of a Transformer. Time2Vec is a model-agnostic temporal encoding, leading towards model-based deep learning \cite{shlezinger2022modelbaseddeeplearning}, possibly offering transparent and interpretable decision-making.

\section{Background}
\subsection{Transformer}

Transformers \cite{Transformer} are state-of-the-art sequence-to-sequence deep learning models that leverage the so-called self-attention mechanism, that allows the model to weigh the importance of each token in a sequence relative to others. By processing all tokens simultaneously, transformers can better understand context across the entire input, and capture long-range dependencies without the sequential limitations of RNNs. Furthermore, transformers show great ability to model interactions in sequential data, and thus they are appealing to time series modeling. Many variants of Transformer have been proposed to address special challenges in time series modeling and have been successfully applied to various time series tasks, such as forecasting, anomaly detection, and classification \cite{wen2023transformers}. Usual transformer architectures involves Encoder and Decoder parts. Here, we focus on the Encoder only, that consists of multiple identical layers, each containing a multi-head self-attention mechanism and a feed-forward network, designed to process and understand input sequences by focusing on relationships within the data.

%The innovation of the Transformer \cite{Transformer} in deep learning has brought great interest recently due to its excellent performances in NLP \cite{devlin-etal-2019-bert}. Transformers have shown ``significantly'' great modeling ability for long-range dependencies and interactions in sequential data and thus are appealing to time series modeling. Many variants of Transformer have been proposed to address special challenges in time series modeling and have been successfully applied to various time series tasks, such as forecasting, anomaly detection, and classification \cite{wen2023transformers}. The Encoder (of the Transformer) is a crucial component designed to process and understand input sequences by focusing on relationships within the data. It consists of multiple identical layers, each containing a multi-head self-attention mechanism and a feed-forward network. The self-attention mechanism allows the Encoder to weigh the importance of each token in a sequence relative to others, enabling it to capture long-range dependencies without the sequential limitations of RNNs. By processing all tokens simultaneously, the Encoder can better understand context across the entire input.

\subsection{Time2Vec}
Time2Vec \cite{Time2vec} is an approach providing a model–agnostic vector representation for time. The input signal is decomposed in the time domain to $k+1$ components of the form
\[
\mathbf{t2v}(\tau)[i] = \begin{cases}
\omega_i \tau + \varphi_i, & \text{if } i = 0, \\
\mathcal{F}(\omega_i \tau + \varphi_i), & \text{if } 1 \le i \le k,
\end{cases}
\]
where $\tau$ denotes the time instance, $\mathcal{F}$ is a periodic activation function, and $\omega_i$ and $\varphi_i$ are learnable parameters. Note that if $\mathcal{F}$ is a sinusoid, then Time2Vec provides a time-invariant decomposition that encodes a temporal signal into a set of frequencies. The model was inspired by parts of positional encoding from \cite{Transformer}, therefore it can be directly combined with the transformer architecture. In this sense, Time2Vec serves as an alternative to positional encoding that can also capture periodic behavior not supported by positional encoding.

\subsection{Correlation Analysis}

% Auto-correlation and cross-correlation are essential tools for examining temporal dependencies within and between time series data. They help identify repeating patterns, trends, and the relationship between different variables over time. Correlation coefficients range from $-1$ to 1 and quantify whether, and how strongly, pairs of variables are related. A positive correlation indicates that as one variable increases, the other tends to increase as well, while a negative correlation indicates an inverse relationship. When two variables are uncorrelated, changes in one variable do not systematically affect the other.

In order to identify repeating patterns, trends, and temporal dependencies within and between time series data, we examined their autocorrelation and cross-correlation. We considered the normalized cross-correlation of time series $x$ and $y$ as a function of lag $k$ of the form
\[
\rho_{xy}(k) = \frac{1}{\sigma_x \sigma_y} \sum_t\left(x_t - \overline{x}\right)\left(y_{t+k} - \overline{y}\right),
\]
where $\overline{x}$, $\overline{y}$, $\sigma_x$ and $\sigma_y$ denotes the means and the standard deviations of $x$ and $y$, respectively. Autocorrelation of $x$ is computed as $\rho_{xx}$, i.e. the cross-correlation with itself. In principle, correlation quantifies the linear relationship between the inputs. The values range between $-1$ and $1$ due to the normalization, where $1$, $0$, and $-1$ indicate perfect correlation, no correlation, and anticorrelation, respectively.

%\subsection{Geometric Mean Not NaN (GMNN)}
\subsection{Feature aggregation}

% Combining multiple data may create NaN values due to the difference in the size of each data. To resolve the obstacle, we use GMNN. GMNN ensures the transformed data will be in between 0 and 1, which is the must-have attribute of the data. GMNN iterates by row and takes the GM of real numbers. In this paper, GMNN performs column-wise, meaning that it only performs on the related columns which means with two tables input of size $m\times n$, the result's size still remains $m\times n$.

We investigated statistical aggregation methods to combine features from multiple time series to a single feature vector, which provides a simplified data representation that still benefits from the variations of multiple sources. Although the input time series are sampled at the same frequency, there are missing data points (treated as NaN), due to the size differences and missing information in the database. In order to resolve this obstacle, we used Geometric Mean Not NaN (GMNN), where we computed the geometric mean of different sources at each time point, ignoring NaN values.
\section{Methodology}
\subsection{Motivation}
\begin{figure}[htbp]
\centerline{\includesvg[width=\linewidth]{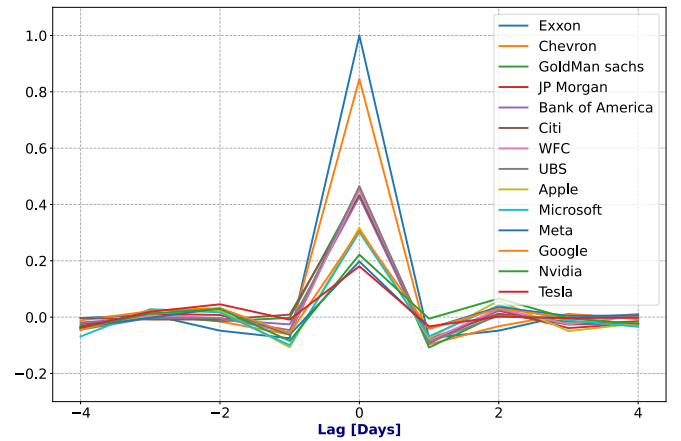}}
\caption{Correlation analysis of stock indices: auto-correlation and cross-correlation of market trends compared to Exxon Mobil, inspired by \cite{Shen2012StockMF}.}
\label{fig:nas_base}
\end{figure}

The performance of a stock market predictor heavily depends on the correlations between historical data for training and the current input for prediction.
%The results show in Fig.~\ref{fig:nas_base} reveals that the base market’s auto-correlation is solely non-zero at the origin. This observation suggests that the daily trend of NASDAQ stock approximates a Markov process \cite{Shen2012StockMF}. Consequently, historical data offers limited insight into its future movements. However, data sources such as S\&P500 are promising features for our approach. 
Motivated by \cite{Shen2012StockMF}, we investigated autocorrelation and cross-correlation of multiple groups of stock indices to reveal short-term temporal dependencies and relations. As an example, Fig.~\ref{fig:nas_base} depicts the daily correlations using Exxon Mobil as base market. Similar to \cite{Shen2012StockMF}, we have found that the autocorrelation of the base market is mostly zero outside the origin, that suggests possible Markovian property. Consequently, we can presume that historical data offers limited insight into its future movements. On the other hand, different data sources such as Chevron provide a promising supporting feature due to the high correlation, and possible actual causation.
Motivated by the strong correlation observed between the selected stock pairs, we propose a novel method to combine multiple stocks into a single input representation. By leveraging cross-correlation relationships, our approach aims to capture the joint dynamics of these correlated stocks, enabling the model to better understand their collective behavior and improve predictive accuracy.

\subsection{Data Acquisition}
\label{sec:data}

\begin{figure}[htbp]
\centerline{\includegraphics[width=\linewidth]{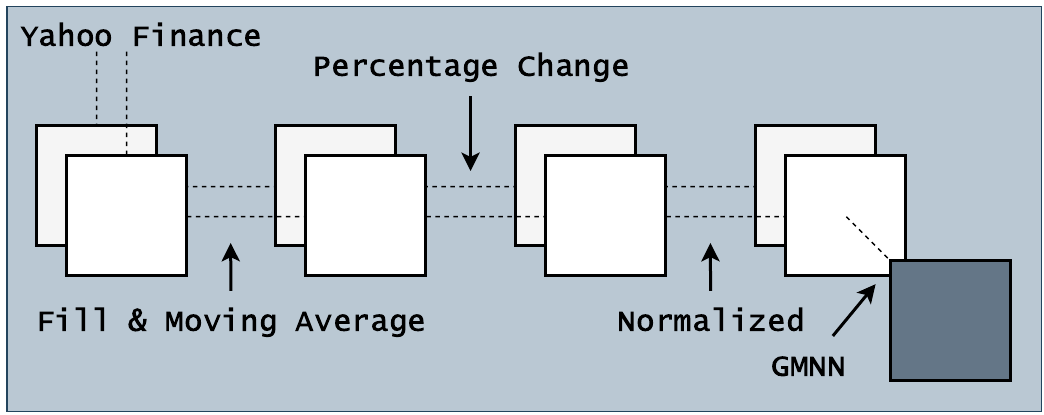}}
\caption{The preprocessing pipeline. Steps: Fill and Moving Average, Percentage Change Computing, Normalization, Aggregation using GMNN.}
\label{fig:pipeline}
\end{figure}

Daily stock datasets are collected from their release date up to the 30th of September, 2024, using Yahoo Finance \cite{yf}. Based on the above motivation, we collected three groupings of two equities whose members significantly correlated to each other, such as \textbf{Group A:} NASDAQ (\texttt{\^}IXIC), S\&P500 (\texttt{\^}GSPC); \textbf{Group B:} Exxon Mobil (XOM), Chevron (CVX); and \textbf{Group C:} Morgan Stanley (MS), Goldman Sachs (GS).

We propose the preprocessing and feature selection pipeline depicted on Fig.~\ref{fig:pipeline}. First, missing data points are filled to create consistent and continuous inputs. Then, moving average is applied with a 14-day period in order to compensate for short-term fluctuations and produce smoother local trends. Next, percentage changes are computed for the next day, normalized between 0 and 1. Finally, once these steps are applied to each stock indices in the group, the processed data of the whole group is aggregated using GMNN to a single feature, that will serve as the final input for the model.
% The preprocessing data pipeline is shown in Fig.~\ref{fig:pipeline}, since moving averages and percentage changes work on continuous inputs, discontinuities would cause inconsistency. To address the obstacle, we filled all the missing dates. To smooth out the ups and downs in prices, we started by averaging the values over a 14-day period, then calculating the percentage change for the next day. Repeat the steps with the other stock indices in the group, then we combine all the processed data in the same group thanks to GMNN, which then will be the final input for the model. 

\begin{figure}[htbp]
\centerline{\includegraphics[width=\linewidth]{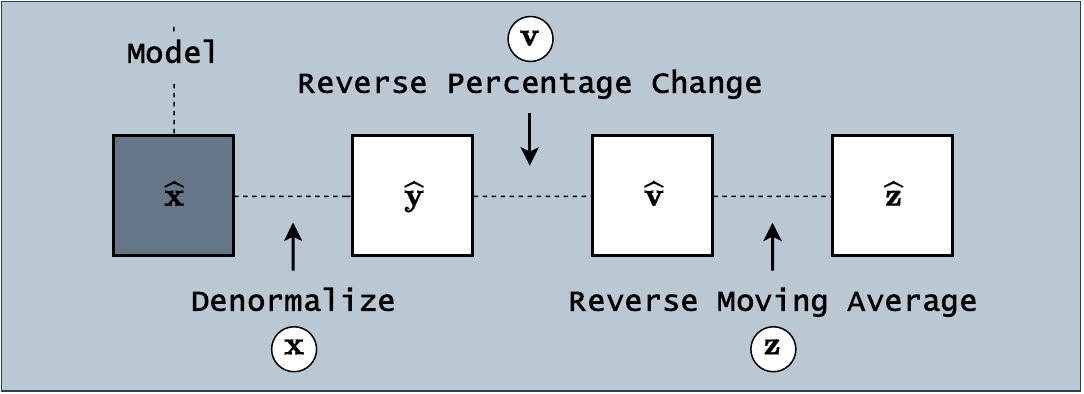}}
\caption{The postprocessing pipeline. Steps: Denormalization, Reverse Percentage Change, Reverse Moving Average. $\mathbf{x}$ is the original normalized values, $\mathbf{\widehat{x}}$ is the predicted normalized values, $\mathbf{\widehat{y}}$ is the predicted percentage change values, $\mathbf{v}$ is the original moving average values, $\mathbf{\widehat{v}}$ is the predicted moving average values, $\mathbf{z}$ is the original close prices, $\mathbf{\widehat{z}}$ is the predicted close prices.}
\label{fig:postprocessing}
\end{figure}

After the predictions, we applied postprocessing to retrieve back the close prices, as of Fig.~\ref{fig:postprocessing}.
% Fig.~\ref{fig:postprocessing} shows the pipeline of retrieving back close prices.
The output from the model is clipped in the range from 0 to 1 during the denormalization step. Then, the reverse percentage change and reverse moving average are computed to get desired close prices.

\subsection{Proposed Model}

\begin{figure}[htbp]
\centerline{\includegraphics[width=\linewidth]{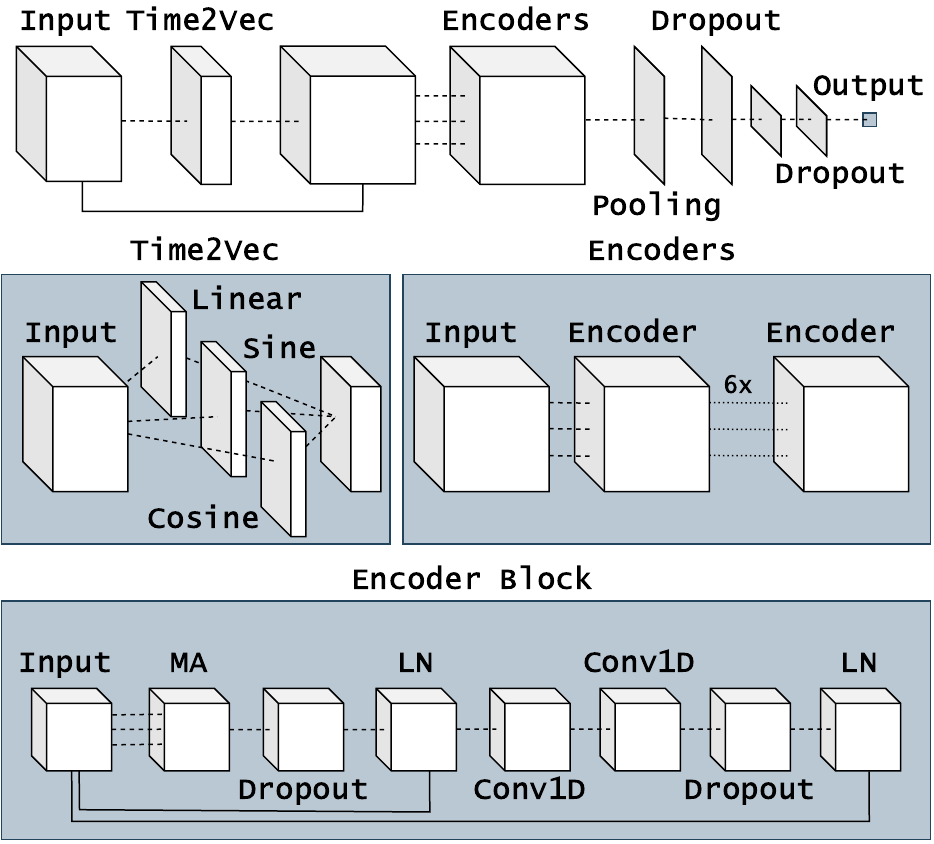}}
\caption{The proposed model and architectures. Proposed model (top), Time2Vec module (left mid), Transformer Encoders module (right mid), Encoder Block (bottom). MA stands for Multi-head Attention, LN stands for Layer Normalization. The solid lines are residual connections.}
\label{fig:model}
\end{figure}

Inspired by the inherent limitations of complex deep learning models, which are frequently trained on historically inconsistent and unreliable datasets, as well as the common data challenges across diverse industries such as technology, finance, energy, and capital markets, this research aims to tackle critical methodological constraints in predictive analytics. We introduce a novel network architecture, presented in Fig.~\ref{fig:model}, that combines the Time2Vec and the Encoder of the Transformer. In detail, Time2Vec encodes positions and captures both linear and periodic behaviors of prices over time; the multiple Encoders perform multi-head self-attention, layer normalization, as well as positional-wise feed-forward; Residual connections are used to improve the behavior of the network which leads to a more valuable and powerful model \cite{resnet}; Pooling reduces the spatial dimensions of the output from the Encoder, in terms of the number of features, while retaining the most important information; Dropout is an indispensable layer to avoid overfitting problem.

% \textcolor{red}
{Our design concepts were the to develop a robust architecture that can utilize the expression and generalization ability of the attention mechanism through Encoders, and to involve Time2Vec as encoding. Our investigations proved that Time2Vec is a fitting encoding scheme for the target application, and provides better ability to capture temporal patterns compared to the traditional approaches, as discussed in Section~\ref{sec:results}.}
\section{Results and Evaluation}
\label{sec:results}

We evaluated multiple models for comparison purposes, where in each case we trained the models on the first 80\% of the data, validated on the next 10\%, and the performance evaluations are made using the last 10\%. Common evaluation metrics are used, such as root mean square error (RMSE), mean square error (MSE), mean absolute percentage error (MAPE), mean absolute error (MAE), R-square (R2).

In Table~\ref{tab:evaluation}, we compared the performance of single-feature models and the proposed multi-feature approach. Here, we used the following notations: models such as NAS, SP, EM, C, MS, and GS are single-feature models, meaning that we only use their target for training; on the other hand, NAS\_SP, EM\_C, and MS\_GS are multi-feature models; they are trained by using the combined dataset (e.g., NASDAQ and S\&P500 for NAS\_SP). To evaluate models in each group, we rely on their target, for instance: NAS is a single-feature model with NASDAQ as a target, NAS\_SP is a multi-feature model with NASDAQ as a target as well. Then, NAS\_SP is compared to NS with respect to the same target, which is NASDAQ in this case.
% Then to know how good NAS\_SP is better than NAS, we will compare them with respect to the same target which is NASDAQ.
By this evaluation, we expect to figure out how well the cross-correlation affects the models' performance. We can conclude that the multi-feature models outperform single-feature models in 2 out of 3 cases, i.e. the feature selection benefits from the proposed feature aggregation. The only exception is the circumstance where the stock ages are young (i.e., their release date is about 10 to 20 years ago) and the cross-correlations are not tight enough.
In addition, we can point out that the cross-correlation of NASDAQ and S\&P500 is approximately 91\%. Between Exxon Mobil and Chevron it is nearly 84\% (see Fig.~\ref{fig:nas_base}), and between Morgan Stanley and Goldman Sachs it is about 81\%.
%(To save spaces for other important information, we present the first figure only). 
Based on those clues, we conclude that the better the link between stocks, the better the multi-feature prediction.

\begin{table}[htbp]
    \centering
    \caption{{{Evaluations by Target of Each Group}}}
    \resizebox{\linewidth}{!}{%
    \begin{tabular}{ccccccc} 
		
    \textbf{Target} & \textbf{Model} & \textbf{RMSE} & \textbf{MSE} & \textbf{MAPE} & \textbf{MAE} & \textbf{R2} \\ 
    \hline \hline \\[-8pt]
    Group A & &&&&&
    \\ 
    \hline \hline \\[-8pt]
    
    \multirow{2}{*}{\textbf{\texttt{\^}IXIC}} &NAS & 0.0291 & 0.0008 & 3.1392 & 0.0187 & 0.8264 \\
    & \multirow{2}{*}{NAS\_SP} & \textbf{0.0248} & \textbf{0.0006} & \textbf{2.6785} & \textbf{0.0161} & \textbf{0.8743} \\ 
    \cline{1-1} \cline{3-7} \\[-8pt]
    \multirow{2}{*}{\textbf{\texttt{\^}GSPC}} & & \textbf{0.0164} & \textbf{0.0003} & \textbf{1.9422} & \textbf{0.0098} & \textbf{0.8766} \\
    &SP & 0.0180 & \textbf{0.0003} & 2.0511 & 0.0104 & 0.8517 \\

    \hline \hline \\[-8pt] 
    Group B & &&&&&
    \\ 
    \hline \hline \\[-8pt]

    \multirow{2}{*}{\textbf{XOM}} &EM & 0.0350 & 0.0012 & 4.3509 & 0.0204 & 0.8163 \\
    &\multirow{2}{*}{EM\_C} & \textbf{0.0333} & \textbf{0.0011} & \textbf{4.0724} & \textbf{0.0200} & \textbf{0.8330} \\ 
    \cline{1-1} \cline{3-7}\\[-8pt]
    \multirow{2}{*}{\textbf{CVX}}& & \textbf{0.0275} & \textbf{0.0008} & \textbf{3.3767} & \textbf{0.0149} & \textbf{0.8241} \\
    &C & 0.0342 & 0.0012 & 3.7204 & 0.0172 & 0.7293   \\

    \hline \hline \\[-8pt]
    Group C & &&&&&
    \\ 
    \hline \hline \\[-8pt]
    
    \multirow{2}{*}{\textbf{MS}} & MS & \textbf{0.0148} & \textbf{0.0002} & \textbf{1.7162} & \textbf{0.0105} & \textbf{0.8655} \\
    &\multirow{2}{*}{MS\_GS} & 0.0166 & 0.0003 & 2.0747 & 0.0128 & 0.8325 \\ \cline{1-1} \cline{3-7}\\[-8pt]
    \multirow{2}{*}{\textbf{GS}} & & 0.0221 & 0.0005 & 2.8797 & 0.0156 & 0.8545\\
    & GS& \textbf{0.0201} & \textbf{0.0004} & \textbf{2.6159} & \textbf{0.0144} & \textbf{0.8797} \\
		
    \end{tabular}
    }
    \label{tab:evaluation}
\end{table}

\begin{table}[htbp]
    \centering
    \caption{{{Comparing to State-of-the-art Models}}}
    \resizebox{\linewidth}{!}{%
    \begin{tabular}
    {clccccc} 
    \textbf{Target} & \textbf{Model} & \textbf{RMSE}  & \textbf{MSE} & \textbf{MAPE}  & \textbf{MAE} & \textbf{R2} \\ \hline \hline \\[-8pt]

     Group A & & & & & & \\ \hline \hline \\[-8pt]
    
    \multirow{6}{*}{\textbf{\texttt{\^}IXIC}}&\textbf{TT2VFin}& \textbf{0.0248} & \textbf{0.0006} & \textbf{2.6785} & \textbf{0.0161} & \textbf{0.8743} \\
    &TT2VFin+M&0.0293 &0.0009 &3.3222 &0.0205 &0.8240  \\
    &TT2VFin+p&0.1773 &0.0314 &27.500 &0.1691 &-- 5.427 \\
    &TT2VFin+p+M& 0.0337 & 0.0011&3.9642 &0.0241 &0.7685 \\
    &Transformer+p&0.0382 &0.0015 &4.4677 &0.0279 &0.7008 \\
    &RNN & 0.0316 & 0.0010 & 3.6215 & 0.0224 & 0.7959 \\
    &LSTM & 0.0272 & 0.0007 &2.9289 &0.0176 & 0.8490 \\

     \hline \\[-8pt]
    
    \multirow{6}{*}{\textbf{\texttt{\^}GSPC}}&\textbf{TT2VFin}& \textbf{0.0164} & \textbf{0.0003} & \textbf{1.9422} & \textbf{0.0098} & \textbf{0.8766} \\
    &TT2VFin+M&0.0180 &\textbf{0.0003} &2.2002 &0.0109 &0.8515 \\
    &TT2VFin+p&0.2265 &0.0513 &42.337 &0.2253 &-- 22.51 \\
    &TT2VFin+p+M&0.0247 &0.0006 &3.6617 &0.0189 &0.7204 \\
    &Transformer+p&0.0207 &0.0004 &2.5934 &0.0132 &0.8024 \\
    &RNN& 0.0195 & 0.0004 & 2.3061 & 0.0116 & 0.8262\\
    &LSTM& 0.0184 & \textbf{0.0003} & 2.3468 & 0.0119 & 0.8448 \\

    \hline \hline \\[-8pt]

     Group B & & & & & & \\ \hline \hline \\[-8pt]

    \multirow{6}{*}{\textbf{XOM}}&\textbf{TT2VFin}& \textbf{0.0333} & \textbf{0.0011} & \textbf{4.0724} & \textbf{0.0200} & \textbf{0.8330} \\
    &TT2VFin+M&0.0460 & 0.0021 & 5.8378 & 0.0303 & 0.6824\\
    &TT2VFin+p&0.0859 & 0.0074 & 11.169 & 0.0619 & -- 0.108 \\
    &TT2VFin+p+M& 0.0870 & 0.0076 & 11.315 & 0.0632 & -- 0.137\\
    &Transformer+p&0.0408 &0.0017 &5.4162 &0.0292 & 0.7495 \\
    &RNN & 0.0351 & 0.0012 & 4.5526 & 0.0239 & 0.8150 \\
    &LSTM & 0.0342 & 0.0012 &4.3279 &0.0224 &0.8248 \\

     \hline \\[-8pt]
    
    \multirow{6}{*}{\textbf{CVX}}&\textbf{TT2VFin}& 0.0275 & \textbf{0.0008} & \textbf{3.3767} & \textbf{0.0149} & 0.8241 \\
    &TT2VFin+M&0.0380 & 0.0014 & 5.2774 & 0.0237 & 0.6661\\
    &TT2VFin+p&0.0713 & 0.0051 & 10.285 & 0.0500 & -- 0.176\\
    &TT2VFin+p+M&0.0700 & 0.0049 & 10.007 & 0.0486 & -- 0.134\\
    &Transformer+p&0.0382 &0.0015 &4.4677 &0.0279 &0.7008 \\
    &RNN& 0.0287 & \textbf{0.0008} & 3.6525 & 0.0172 & 0.8093 \\
    &LSTM& \textbf{0.0274} & \textbf{0.0008} & 3.4605 & 0.0158 & \textbf{0.8261} \\ \hline \hline \\[-8pt]

     Group C & & & & & & \\ \hline \hline \\[-8pt]

    \multirow{6}{*}{\textbf{MS}}&\textbf{TT2VFin}& \textbf{0.0166} & \textbf{0.0003} & \textbf{2.0747} & \textbf{0.0128} & \textbf{0.8325}  \\
    &TT2VFin+M&0.0174&\textbf{0.0003}&2.1018&0.0130&0.8152\\
    &TT2VFin+p&0.0495&0.0024&6.4577&0.0409&-- 0.494\\
    &TT2VFin+p+M&0.0484&0.0023&6.3163&0.0399&-- 0.430\\
    &Transformer+p&0.0177 &\textbf{0.0003} &2.0849 & 0.0130 & 0.8097 \\
    &RNN &  0.0258 & 0.0007 & 3.6253 & 0.0227  & 0.5933\\
    &LSTM & 0.0186 & \textbf{0.0003} & 2.4826 & 0.0149 & 0.7891 \\

     \hline \\[-8pt]
    
    \multirow{6}{*}{\textbf{GS}}& {TT2VFin}&{0.0221} & \textbf{0.0005} & {2.8797} & {0.0156} & {0.8545} \\
    &TT2VFin+M&0.0236&0.0006&3.2474&0.0176&0.8343 \\
    &TT2VFin+p&0.0661&0.0044&10.086&0.0521&-- 0.304\\
    &TT2VFin+p+M&0.0671&0.0045&10.265&0.0529&-- 0.343\\
    &\textbf{Transformer+p}&\textbf{0.0214} &\textbf{0.0005} &\textbf{2.8383} & \textbf{0.0154} & \textbf{0.8621} \\
    &RNN&  0.0246 & 0.0006 & 3.4026 & 0.0192 & 0.8200 \\
    &LSTM& 0.0280 & 0.0008 & 4.3016 & 0.0229 & 0.7669 \\ \\
    \end{tabular}%
    }
    \raggedright
    \footnotesize In this experiment, TT2VFin and its variations use 143K parameters, {Transformer+p uses 173K parameters,} RNN uses 559K parameters, and LSTM uses 355K parameters.
    \label{tab:compare}
\end{table}

\begin{figure*}[htbp]
\centering
% {\includesvg[width=\linewidth]{figs/nasdaq_stock_price_plot1.svg}}
% \hfill
{\includesvg[width=\linewidth]{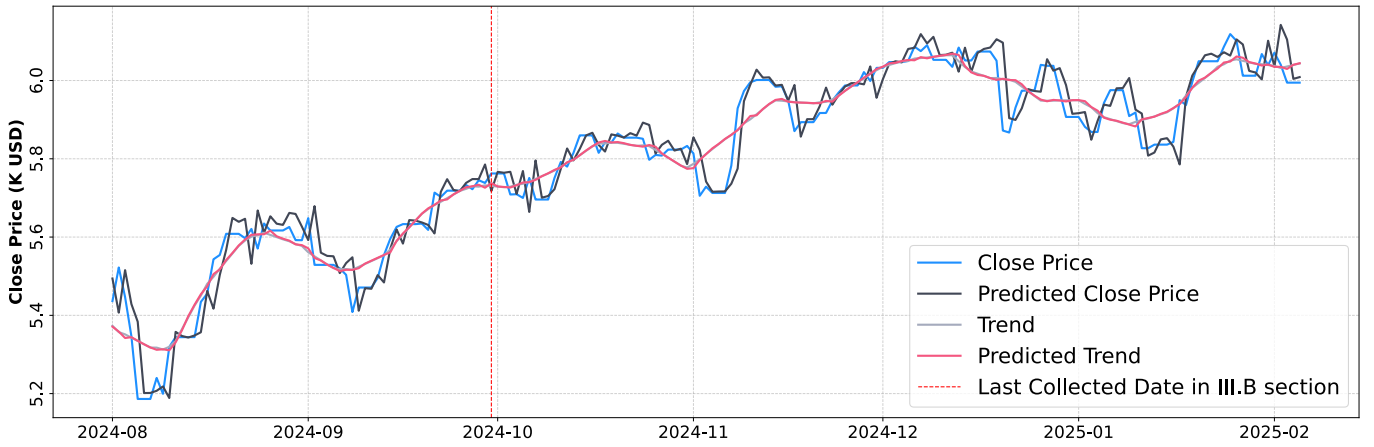}}
% \caption{The predicted close prices. NASDAQ (top) and S\&P500 (bottom) predicted close prices and orginal close prices are displayed in the same interval of 6-month (from 2024-08-01 to 2025-02-05) together with their trends and predicted trends. The last date of data was used for training is 2024-09-30.}
\caption{The predicted close prices. S\&P500 predicted close prices and orginal close prices are displayed in the same interval of 6-month (from 2024-08-01 to 2025-02-05) together with its trend and predicted trend. {The red line demonstrates the last collected date we did in Section~\ref{sec:data}. Note that a similar pattern of close prices and trends are attained by the NASDAQ stock.}}
\label{fig:close_price}
\end{figure*}

In Table~\ref{tab:compare}, we compared our model with different encoding techniques, and state-of-the-art deep learning predictor models, such as RNN and LSTM. Here, \textit{TT2VFin} refers to our proposed multi-features model, \textit{TT2VFin+p} adds positional encoding layer, \textit{TT2VFin+M} uses masking in the attention mechanism, \textit{TT2VFin+p+M} uses both positional encoding and masking. The results show that \textit{TT2VFin} outperforms its variations,
%Applying a mask does not significantly affect \textit{TT2VFin} performance; however, the positional encoding layer confuses the model and produces poor results.
% \textcolor{red}
{and \textit{Transformer+p} is a Transformer Encoder using positional encoding instead of Time2Vec. As a more detailed analysis, we can see that applying a mask does not significantly affect \textit{TT2VFin}; however, the incorporation of positional encoding to the proposed model reduces its performance. The likely reason behind this behavior is that Time2Vec itself acts as an encoding scheme, and the combination of the two encodings confuses the model and leads to suboptimal performance. We also see that \textit{TT2VFin} outperforms \textit{Transformer+p} in 5 out of 6 cases and has comparable performance in one case, that proves that Time2Vec is more efficient for the time-series domain than the original positional encoding. Regarding the baseline deep learning models, the} proposed model outperforms RNN despite utilizing about three times fewer parameters as RNN. LSTM demonstrates its power in financial prediction; nonetheless, there is still a performance difference between \textit{TT2VFin} and LSTM. Statistically, \textit{TT2VFin} outperforms LSTM by around 9.25\% $\pm$ 0.63\% in Group~A (averaging the metrics). We achieved the same results in other groups, where our proposed model is dominating over others. \textit{TT2VFin+p} and \textit{TT2VFin+M} have no change in performance, but \textit{TT2VFin+p+M} cannot maintain its previous performance. In general, our proposed model uses 2 to 3 times fewer parameters than others but has roughly 14.23\% greater performance (by averaging all the metrics differences between base and RNN and LSTM). %\textcolor{red}{In comparison with the Transformer Encoder using positional encoding instead of Time2Vec, we see that \textit{TT2VFin} outperforms \textit{Transformer+p} in 5 out of 6 cases. Moreover, in the losing case, the difference is not too big that showing the efficiency of Time2Vec instead of original positional encoding for time-series domain.} %Applying positional encoding confuses \textit{TT2VFin} rather than improving it. Applying the mask does not significantly degrade performance, but it also does not increase it. %% already mentioned!

Additionally, we also investigated the performance of real close price predictions, visualized on Fig.~\ref{fig:close_price}.
%To satisfy the needs of visualizing the close prices prediction, we present the Fig.~\ref{fig:close_price}, where we show how good our model perform in predicting the real prices.
The model preserves its performance when facing with new data, which proves its generalization ability, specially in these days when the market is extremely unstable due to multiple factors.

\section{Conclusion}
In this paper, we investigated the application of deep learning for stock price prediction, which is a challenging time series forecasting problem in the financial industry.
{Notably, we represent a novel methodology for combining correlation features, aiming to better capture inter-stock relationships and temporal dynamics}. We studied different markets by selecting correlation features to enhance the accuracy in predicting multiple stock prices and developed a novel neural network architecture that combines the Encoder of the Transformer with the Time2Vec model. Through careful testing and analysis, we have found that our model performs exceptionally well, better than standard models such as LSTM and RNN, and encoding methods like positional encoding from Transformer. In most cases, our multi-feature proposed model is better than a single-feature model, that verifies the importance and potential of feature selection and aggregation. In conclusion, our study demonstrated that techniques like Time2Vec and Encoder are promising future approaches combined with multi-source feature selection. We also believe that our approach can be successfully adopted and generalized to related financial and other time series prediction problems, and can achieve even better results for more refined data sources (e.g., having the intra-day historical data).

%This success is because our model looks at the bigger picture, using a mix of different features that work together seamlessly, setting a new standard for how good predictions can be in this field. These findings show that using advanced methods for predicting can make a big difference. They also remind us how important to think about how complicated our model is and which features we use. By paying attention to these details, our study shows a way to make predictions that work well in real-life situations. In the future, people can try more variations of Time2Vec and Encoder to expect a higher accuracy. We believe that scaling up the dataset (e.g., having the intra-day historical data) will allow the model to perform much better.

\section*{Acknowledgment}

We thank our friend, Ho Minh Tri, for his resources assistance during the model training process.

\bibliographystyle{./IEEEtran}
\bibliography{./IEEEabrv,./mybib}

\end{document}